\documentclass[letterpaper, 10 pt, conference]{ieeeconf}  %

\usepackage[linesnumbered, ruled, vlined]{algorithm2e}  %
\usepackage{hyperref} %
\usepackage{graphicx} %
\usepackage{multirow}
\usepackage{booktabs}
\usepackage{amsmath}
\usepackage{setspace}
\usepackage{enumerate}
\usepackage{threeparttable} %
\hypersetup{
    colorlinks=true,
    linkcolor=blue,
    urlcolor=blue,
    citecolor=blue,
    allbordercolors={0 0 0}, %
    pdfborderstyle={},       %
}

\graphicspath{{images/}} %

\IEEEoverridecommandlockouts                              %

\title{\LARGE \bf
Can a Robot Walk the Robotic Dog: Triple-Zero Collaborative Navigation for Heterogeneous Multi-Agent Systems
}

\author{Yaxuan Wang$^{\dagger 1}$, Yifan Xiang$^{\dagger 2}$ Ke Li$^{3}$, Xun Zhang$^{2}$, Bowen Ye$^{2}$, \\
Zhuochen Fan$^{4,2}$, Fei Wei$^{5}$ and Tong Yang$^{*2}$%
\thanks{$^{1}$Yuanpei College, Peking University}%
\thanks{$^{2}$School of Computer Science, Peking University}%
\thanks{$^{3}$School of Computer Science, Beijing University of Posts and Telecommunications}%
\thanks{$^{4}$Pengcheng Laboratory}%
\thanks{$^{5}$Beijing Jinruyi Large Model Technology Co., Ltd.}%
\thanks{$^{*}$Corresponding author: Tong Yang. Email: yangtong@pku.edu.cn.}%
\thanks{$^{\dagger}$These authors contributed equally to this work.}%
\thanks{This work was supported by the National Key Research and Development Program of China under Grant No. 2024YFB2906603, and in part by the National Natural Science Foundation of China (NSFC) (No. 62372009). The authors would like to thank Beijing Jinruyi Large Model Technology Co., Ltd. for their strong support and close collaboration.The authors also thank Yuanpei College, Peking University, for its support.}
}

\begin{document}

\maketitle
\thispagestyle{empty}
\pagestyle{empty}

\begin{abstract}

We present Triple Zero Path Planning (TZPP), a collaborative framework for heterogeneous multi-robot systems that requires zero training, zero prior knowledge, and zero simulation. TZPP employs a coordinator--explorer architecture: a humanoid robot handles task coordination, while a quadruped robot explores and identifies feasible paths using guidance from a multimodal large language model. We implement TZPP on Unitree G1 and Go2 robots and evaluate it across diverse indoor and outdoor environments, including obstacle-rich and landmark-sparse settings. Experiments show that TZPP achieves robust, human-comparable efficiency and strong adaptability to unseen scenarios. By eliminating reliance on training and simulation, TZPP offers a practical path toward real-world deployment of heterogeneous robot cooperation.
Our code and video are provided at: \hyperlink{https://github.com/triple-zeropp/Triple-zero-robot-agent}{https://github.com/triple-zeropp/Triple-zero-robot-agent}.

\end{abstract}

\section{INTRODUCTION}

Multi-robot systems (MRS) have shown significant potential in applications ranging from logistics to disaster response. In particular, heterogeneous MRS can leverage the complementary strengths of different platforms (e.g., the mobility of quadrupeds and the manipulation capabilities of humanoids) to perform complex tasks. Among these tasks, path planning is a key challenge for such collaboration. Currently, an emerging direction to address this problem is multi-robot collaboration and path planning based on large language models (LLMs) \cite{11, 20, 21, 26, 27}.

Existing studies on multi-agent collaborative path planning mainly fall into three categories: (1) learning-based approaches that rely on large-scale training or fine-tuning (e.g., Graph-Based \cite{20}, TaskExp \cite{9}), which can tackle novel problems in unfamiliar environments to some extent but incur high training and deployment costs; (2) methods dependent on prior maps or scene modeling (e.g., ZeroCAP \cite{15}, COHERENT \cite{28}), which perform well in known environments but tend to generalize poorly and fail in real, dynamic settings; and (3) simulation-driven approaches (e.g., SIGMA~\cite{29}, Graph-Based \cite{20}), which train and validate mainly in virtual environments and thus heavily depend on the quality of the simulation--if there is a large gap or poor fidelity between simulation and reality, the real-world performance can be severely affected. A summary of how representative prior works align with the three aforementioned criteria is provided in Table~\ref{tab:related_work}.

\begin{table}
\centering
\caption{Comparison Against the Triple Zero Criteria}
\small
\begin{tabular}{@{}lccc@{}}
\toprule
Method & \shortstack{No \\ Training} & \shortstack{No Prior \\ Knowledge} & \shortstack{No \\ Simulation} \\
\midrule
Ours & $\surd$ & $\surd$ & $\surd$ \\
ZeroCAP \cite{15} & $\surd$ & $\times$ & $\times$ \\
TaskExp \cite{9} & $\times$ & $\surd$ & $\times$ \\
SIGMA \cite{29} & $\times$ & $\times$ & $\times$ \\
Hybrid \cite{6} & $\surd$ & $\times$ & $\surd$ \\
COHERENT \cite{28} & $\surd$ & $\times$ & $\surd$ \\
Graph-Based \cite{20} & $\times$ & $\times$ & $\times$ \\
MIM \cite{7} & $\surd$ & $\times$ & $\surd$ \\
\bottomrule
\end{tabular}
\label{tab:related_work}
\vspace{-0.5cm}
\end{table}

To address these limitations, this paper proposes a novel paradigm for multi-robot path planning---Triple Zero Path Planning (TZPP)---designed to support heterogeneous agents in autonomously exploring and reaching targets in unknown, complex environments. Taking the collaboration between a humanoid agent and a quadruped agent as an example, we construct a coordinator-explorer architecture: the humanoid handles high-level task coordination and navigation, while the quadruped undertakes environment exploration and feasible path identification. We deploy this framework in real-world environments, where the G1-Go2 multi-agent system achieves human-level collaborative efficiency across multiple tasks, stably accomplishing complex navigation missions and effectively emulating human exploration and decision-making behaviors. These results validate the soundness and innovation of the paradigm’s architecture and mechanisms. Overall, the main contributions of this paper are as follows:

\begin{itemize}

\item We propose the first heterogeneous agent path planning paradigm satisfying the ``Triple Zero'' constraints---zero training, zero prior knowledge, and zero simulation dependency. Without additional training or prior conditions, the approach enables multi-robot systems to autonomously explore and plan paths in real, dynamic, and human-free environments, reducing application costs and enhancing system robustness.

\item We design and validate a heterogeneous robot collaboration mechanism in the real-world. We test the path planning paradigm in various settings, including open spaces, obstacle-rich environments, and both indoor and outdoor scenarios. The result demonstrate human-level performance, confirming the method’s practicality and scalability.

\end{itemize}

TZPP lays the groundwork for future LLM-based multi-robot systems in real-world applications. It not only accelerates the transition of multi-robot path planning from simulation to reality but also offers new insights and tools for heterogeneous robot collaboration. We have released part of our data and experimental benchmarks to facilitate further research.

\section{RELATED WORKS}

For tasks that involve exploration and navigation in unknown environments, common approaches include pretraining to enhance generalization in unfamiliar settings \cite{11, 9, 10, 12, 14}, as well as reinforcement learning methods that allow robots to learn efficient navigation strategies through interaction with the environment \cite{8}, including the TaskExp method \cite{9} and diffusion-based approaches \cite{10}. These methods demonstrate better adaptability compared to traditional map-dependent approaches \cite{1, 2, 3, 4} but rely heavily on large amounts of high-quality data for pre-training. Furthermore, due to the scarcity of real-world scene data, most research uses simulated environments \cite{11, 21, 9}. However, studies trained and tested in simulated environments often struggle to translate into practical applications due to the gap between realistic and idealized settings.

\subsection{Collaboration of Heterogeneous Agents}

Homogeneous multiagent collaboration architectures \cite{15, 16, 17, 18, 19} are relatively simple in structural design and control logic, making them easy to implement and scale. However, a lack of appropriate labor management can lead to problems such as planning confusion and resource wastage. In contrast, heterogeneous multi-agent collaboration systems consist of robots with different functionalities or morphologies. They improve execution efficiency through task allocation and specialized design, enabling more rational management over tasks such as navigation and transportation. Representative work includes the GATAR model and perception sharing among heterogeneous robots \cite{11, 20, 21}, which facilitate collaborative decision making among multiple agents and demonstrate favorable scalability and environmental adaptability.

\subsection{Integration of VLMs and Robotics}

Traditional robotic systems depend on predefined rules and specialized perception modules \cite{22, 23}. While these methods perform reliably in structured environments, their generalization capability is limited when faced with open-ended instructions and dynamic environments. Robotic systems incorporating Vision-Language Models (VLMs) significantly enhance the comprehension of natural language instructions and environmental adaptability through multimodal fusion and semantic reasoning \cite{5}. This enables zero-shot generalization to unseen instructions \cite{24, 13, 25}, thereby supporting more intelligent decision-making.

\section{Our TZPP System}

\subsection{Overview}

This study focuses on the path planning problem of heterogeneous multi-agent systems, aiming to leverage the complementary strengths of different agents through division of labor and collaboration to achieve efficient exploration of complex real-world environments. The research covers a variety of representative scenarios (including both indoor and outdoor settings) and diverse terrains (such as slopes and staircases).

The notations and variables used in the following sections are summarized in Table~\ref{tab:code_var}.

In this work, we take a humanoid agent and a quadruped agent as a typical interaction pair: the humanoid serves as a valuable but relatively less mobile ``core'' agent, while the quadruped acts as a lower-value but more mobile auxiliary agent. The objective is to guide the core agent, with the collaborative support of the auxiliary agent, to explore unknown environments and reach designated target locations $L_T$.

It's noteworthy that complex real-world environments pose significant challenges for path planning and task recognition in agent systems. In scenarios lacking salient landmarks, agents struggle to obtain effective localization information, leading to unreliable position estimates, reducing the accuracy of action decisions. What's more, in environments where direct access to the target is not possible, agents often need to detour for the goal.

\begin{figure*}[h]
    \includegraphics[width=1\linewidth]{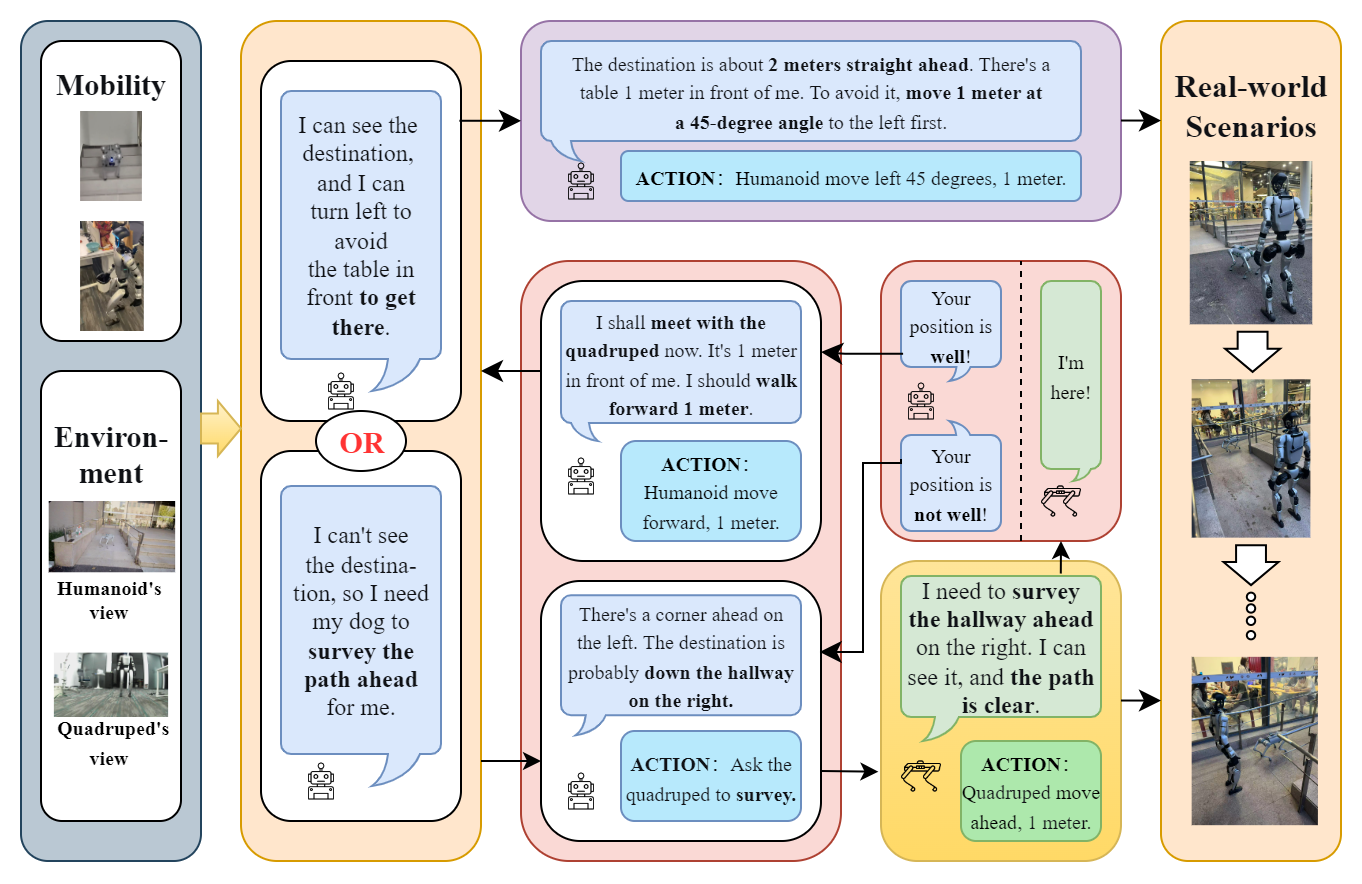}
    \caption{\textit{System pipeline of the proposed multi-agent collaboration framework.} Given observations from both the humanoid and quadruped perspectives, the system performs reasoning to determine whether the target is directly reachable. If the destination is visible, the humanoid plans and executes obstacle-aware navigation. Otherwise, the humanoid delegates exploration to the quadruped, which surveys the environment and provides feedback on path feasibility. Based on this feedback, the humanoid updates its plan and continues navigation. This iterative collaboration enables robust decision-making and efficient task completion in complex real-world scenarios.\protect\footnotemark}
    \label{fig:sys_pipeline}
    \vspace{-0.2cm}
\end{figure*}

To address these challenges, this study adopts a coordinator-explorer architecture design. Specifically, the humanoid, as the representative of the core agent, is responsible for task coordination and reaching the designated location, while the quadruped, as the auxiliary agent, undertakes environment exploration and the identification of feasible paths or intermediate waypoints. A typical pipeline of the design is shown in Fig. \ref{fig:sys_pipeline}.

\begin{table}[h]
\caption{Description of Variables}
    \label{tab:code_var}
    \centering
    \resizebox{0.5\textwidth}{!}{
    \begin{tabular}{@{}clc@{}}
        \toprule
        \textbf{Variable} & \textbf{Description} \\
        \midrule
        $L_T$        & High-level task target location (natural language) \\
        $L_W$        & Quadruped waypoint location (natural language) \\
        $C$          & Inter-agent interaction context \\
        $I_B$        & Humanoid's forward perceptual data \\
        $I_D$        & Quadruped's forward perceptual data \\
        $d_{dt}$     & Euclidean distance to target \\
        $X$          & Landmark-sparse mode \\
        $Y$          & Obstacle-rich mode \\
        $R_{scan}$   & Search half-angle \\
        \bottomrule
    \end{tabular}
}
\end{table}

\subsection{Humanoid Pipeline}

\footnotetext{Robot icons created by Good Ware, robot dog icons created by Izwar Muis - Flaticon}

\begin{figure}
    \includegraphics[width=1\linewidth]{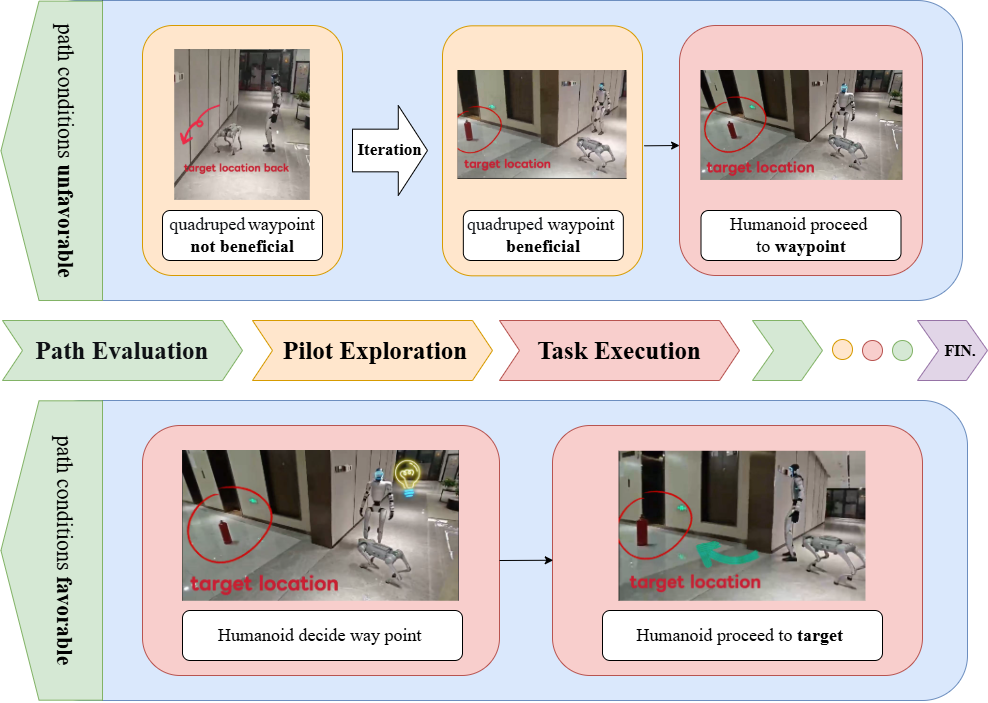}
    \caption{\textit{Iterative decision-making logic of the humanoid agent.} The humanoid evaluates path conditions and adaptively determines whether to proceed directly or request assistance from the quadruped. When path conditions are unfavorable, the humanoid iteratively leverages quadruped-guided exploration to identify beneficial waypoints, refining its navigation plan through feedback. Once a feasible waypoint is established, the humanoid proceeds toward it. In contrast, when path conditions are favorable, the humanoid directly selects a waypoint and navigates to the target without additional exploration. This iterative strategy enables robust and efficient navigation under varying environmental conditions.}
    \label{fig:humanoid_logic}
\end{figure}

The humanoid manages high-level coordination through an iterative cycle of \textbf{path evaluation, pilot exploration and task execution}, as illustrated in Fig. \ref{fig:humanoid_logic}. In the path-evaluation stage, the agent assesses the feasibility of reaching the global target $L_T$ based on its current perceptual input $I_B$, including target visibility, obstacle distribution, and estimated path accessibility. If a feasible path is inferred (e.g., the target is visible and no blocking structure is detected), the system directly proceeds to task execution. Otherwise, it triggers pilot exploration by assigning intermediate waypoint(s) to the quadruped for environment probing.

During the pilot-exploration stage, the humanoid incrementally receives perceptual feedback $I_D$ from the quadruped and evaluates candidate waypoints based on their contribution to improving target visibility or path feasibility. Only waypoints that provide informative or progress-enabling observations are incorporated into the navigation plan, while uninformative exploration results are discarded.

In the task-execution stage, the humanoid executes motion commands based on the updated environmental understanding, continuously re-evaluating feasibility as new observations become available. This closed-loop process allows the agent to progressively refine its path and approach the target location until successful completion.

\subsection{Quadruped Pipeline}

The quadruped agent is responsible for environmental exploration and feasible path identification. We designed two different exploration strategies, \textit{Mode X} and \textit{Mode Y}, to help the quadruped to better accommodate landmark-sparse and obstacle-rich environments respectively. Its logical workflow, as shown in Algorithm \ref{alg:quadruped}, proceeds as follows. First, the agent performs pattern matching based on environmental information to determine whether the current situation corresponds to a obstacle-rich environment. It then enters a cyclic process consisting of four phases: \textbf{environment detection, task execution, target detection and environment exploration}.

In the environment-detection phase, the quadruped performs omnidirectional rotational scanning to assess the visibility of the assigned waypoint. If the waypoint is not visible, it immediately terminates the task and reports infeasibility to the humanoid agent. If the waypoint is visible, it proceeds to the task-execution phase, moves to the waypoint, returns environmental perception information to the humanoid, and enters the target-detection phase.

\begin{singlespacing}
\begin{algorithm}
\SetAlgoLined
\SetAlgoNlRelativeSize{0}
\caption{Quadruped Robot Pipeline}\label{alg:quadruped}
\KwIn{Humanoid's target location $L_T$, Quadruped's waypoint location $L_W$}
\KwOut{Reach waypoint successful/failed}
\eIf{$\mathrm{EnvAllReachable}(I_D)$}{
    mode $\gets X$
}
{
    mode $\gets Y$
}
\For{360 degrees rotation}{
    \If{$\mathrm{InspectFor}(L_W, I_D)$ = false}{
        \Return false
    }
}
\While{true}{
    $\mathrm{MoveTo}(L_W, I_D)$\;
    \For{360 degrees rotation}{
        \If{$\mathrm{InspectFor}(L_T, I_D)$ = true $\mathbf{and}$ $\mathrm{PathIdeal}(L_T, I_D)$ = true}{
            \Return true
        }
    }
    \If{$mode \mathbf{\ is\ } Y$}{
        \For{-$R_{scan}$ to +$R_{scan}$  degrees rotation}{
            \If{$\mathrm{InspectFor}(\text{`passage'}, I_D)$ = true}{
                \Return true
            }
        }
    }
    \Return false
}
\end{algorithm}
\end{singlespacing}

In the target-detection phase, the quadruped performs full-range scanning to detect the visibility of the high-level task target and assess path feasibility. If the target is visible and a feasible path exists, it sends a exploration successful signal and current environmental information to the humanoid. If the target is not in sight, or is visible but unreachable from the current position while the system is operating under a landmark-sparse mode, the quadruped terminates the task and reports current environmental information. In a non-globally accessible mode, the agent additionally performs a locally bounded, angle-restricted scan to probe for the existence of potential passages or corridors.

\subsection{Adaptive Mode X/Y Strategy}

To ensure robust navigation across heterogeneous environments, the robot employs an adaptive switching mechanism between two operational modes: Mode X (landmark-sparse) and Mode Y (obstacle-dense). Mode selection is governed by an environmental accessibility assessment derived from the current perceptual observation $I_D$.

Mode X is activated in regions characterized by high global reachability but a dearth of salient features. In this state, the system prioritizes extensive repositioning and $360^{\circ}$ panoramic scanning. This strategy maximizes target visibility and mitigates redundant exploration resulting from perceptual aliasing in feature-poor environments.

Mode Y is triggered when topographical constraints or high obstacle density impede direct global navigation. Beyond standard waypoint execution, the robot performs constrained scanning within a localized search half-angle $R_{scan}$ to identify traversable corridors or detours. This allows for calculated deviations from the goal-directed path to circumvent obstacles in complex spaces.

This dual-mode framework enables the system to autonomously adapt its exploration behavior to the environmental structure, eliminating the need for environment-specific parameter tuning.

\section{EXPERIMENT}

This experiment aims to address the following questions:

\begin{itemize}
\item To what extent can our TZPP paradigm solve the problem of collaborative exploration in complex environments by heterogeneous multi-agent systems?
\item How capable are our TZPP and its two integrated versions (X and Y) when handling scenes with missing landmarks and indirectly navigable scenarios?
\item What is the application prospect of our heterogeneous multi-agent combination in path planning problems?
\end{itemize}

\subsection{Experimental Setup}

The proposed TZPP method is evaluated across multiple structurally complex, previously unseen physical environments to assess navigation performance in scenarios where the target is non-line-of-sight (NLOS) from the agent’s initial position. The experimental framework utilizes the Unitree G1 Edu humanoid and Unitree Go2 Edu quadruped platforms. The perception and decision-making architecture is powered by the Doubao-vision-3.6 large vision-language model. All trials were conducted in real-world settings without prior simulation or environment-specific fine-tuning to ensure the validity of the system's zero-shot generalization capabilities.

We evaluate the interaction paradigm across six dimensions using 16 metrics. \textbf{Global Task Efficiency (Dimension 1)} is quantified by the total completion time $TIME$, the humanoid’s cumulative travel distance $D$, and its total rotation angle $R$. \textbf{Path Planning Fidelity (Dimension 2)} is assessed via the task completion rate $CR$; the path score $PS$, defined as $100 \times (L_{\mathrm{optimal}}/L_{\mathrm{actual}})$, where $L_\mathrm{optimal}$ and $L_\mathrm{actual}$ represent the optimal and actual path lengths, respectively; and the root mean square error ($RMSE$) of the humanoid’s vertical path deviation:

\begin{equation}
    RMSE = \sqrt{\frac{1}{n} \sum_{j=1}^{n} \min_{i} \|P_j - p(P_j, L_i)\|^2}
\end{equation}

\begin{table}[t]
\centering
\caption{Task finish comparison of TZPP vs. human baseline}
\resizebox{0.5\textwidth}{!}{
\begin{tabular}{c ccc ccc}
\toprule
\multirow{2}{*}{} & \multicolumn{3}{c}{TZPP System} & \multicolumn{3}{c}{Human baseline} \\ 
\cmidrule(lr){2-4} \cmidrule(lr){5-7}
 & $TIME$$\downarrow$ (s) & $D$$\downarrow$ (m) & $R$$\downarrow$ (rad) & $TIME$$\downarrow$ (s) & $D$$\downarrow$ (m) & $R$$\downarrow$ (rad) \\
\midrule
scene 1 & 64.00 & 4.00 & 1.57 & 53.30 & 3.80 & 1.57 \\
scene 2 & 18.22 & 2.60 & 1.05 & 17.47 & 2.57 & 0.79 \\ 
scene 3 & 28.58 & 4.55 & 2.83 & 28.01 & 4.53 & 2.09 \\ 
scene 4 & 120.58 & 6.80 & 4.71 & 80.00 & 6.47 & 4.71 \\
scene 5 & 154.89 & 14.60 & 11.78 & 94.21 & 13.83 & 4.29 \\ 
\bottomrule
\end{tabular}
}
\label{tab:task_completion_vs_human}
\end{table}

\begin{table}[t]
\centering
\caption{Task planning comparison of G1-Go2 system vs. human baseline}
\small
\begin{tabular}{c cc cc}
\toprule
\multirow{2}{*}{} & \multicolumn{2}{c}{G1-Go2 System} & \multicolumn{2}{c}{Human baseline} \\ 
\cmidrule(lr){2-3} \cmidrule(lr){4-5}
 & $PS$$\uparrow$ & $RMSE$$\downarrow$ & $PS$$\uparrow$ & $RMSE$$\downarrow$ \\ 
\midrule
scene 1 & 68.18 & 85.99 & 82.19 & 56.05 \\
scene 2 & 98.08 & 14.42 & 99.23 & 6.24 \\ 
scene 3 & 96.74 & 40.75 & 97.13 & 21.21 \\ 
scene 4 & 92.35 & 120.91 & 97.06 & 58.52\\
scene 5 & 88.18 & 56.54 & 93.13 & 33.22 \\ 
\bottomrule
\end{tabular}
\label{tab:planning_vs_human}
\end{table}

\textbf{Autonomous Exploration (Dimension 3)} and \textbf{Multi-Agent Coordination (Dimension 4)} evaluate the quadruped’s utility. These include the key point discovery count $N_K$; the effective exploration rate $EER$; the scouting frequency $N_E$, which is the total number of instances where the quadruped performs reconnaissance for the humanoid agent; the humanoid movement count ($N_{\mathrm{move}}$); and the guidance efficiency coefficient $V_{GE} = D_h / D_q$, where $D_h$ and $D_q$ represent the cumulative distances traveled by the humanoid and quadruped, respectively. A $V_{GE}$ value approaching unity from the left signifies optimal coordination. \textbf{Environmental Robustness (Dimension 5)} evaluates adaptability in feature-poor scenes via the quadruped's command compliance rate $CCR_q$, revisit counts for both agents $N_{\mathrm{rev}}^h, N_{\mathrm{rev}}^q$, and the quadruped’s redundant rotation count $N_{\mathrm{rot}}^q$.

For \textbf{Constrained Navigation (Dimension 6)}, we introduce the obstacle avoidance coefficient $V_{\mathrm{avoid}}$. Let $P(t)$ be a point on the trajectory and $Q(t) = \text{argmin}_{X \in \mathcal{O}} \|P(t) - X\|$ be its projection onto the obstacle surface $\mathcal{O}$. The avoidance magnitude is defined by the arc length $L_{\mathrm{avoid}}$ of the trajectory formed by $Q(t)$:

\begin{equation}
    L_{\mathrm{avoid}} = \mathrm{ArcLength} \Big( \{ Q(t) \, | \, t \in [0, 1] \} \Big)
\end{equation}
\begin{equation}
    V_{\mathrm{avoid}} = \frac{L_{\mathrm{avoid}}^{\mathrm{(actual)}}}{L_{\mathrm{avoid}}^{\mathrm{(optimal)}}}
\end{equation}

The evaluation encompasses five distinct real-world scenarios: an L-turn sofa search (Scene 1); a unilateral-access narrow pillar (Scene 2); a bilateral-passable pillar (Scene 3); a Z-turn fire extinguisher localization (Scene 4); and a ramp-mediated detour to bypass structural steps (Scene 5).

Unless otherwise specified, the following default parameter settings are employed across all experimental trials: maximum displacement and rotation of agents per turn: $d_{\mathrm{max}} = 2$ m and $R_{\mathrm{max}} = \pi/2$ rad respectively; target achievement threshold $d_{\mathrm{achieve}} = 0.5$ m; localized search half-angle $R_{\mathrm{scan}} = \pi/2$ rad.

\subsection{Comparative Evaluation of TZPP and Human Operators in Collaborative Path Planning}

\begin{table}[t!]
\centering
\caption{Quadrupedal Agent exploration comparison of \\ G1-Go2 system vs. human baseline}
\small
\begin{tabular}{c cc cc}
\toprule
\multirow{2}{*}{} & \multicolumn{2}{c}{G1-Go2 System} & \multicolumn{2}{c}{Human baseline} \\ 
\cmidrule(lr){2-3} \cmidrule(lr){4-5}
 & $N_K$$\uparrow$ & $EER$$\uparrow$ & $N_K$$\uparrow$ & $EER$$\uparrow$ \\ 
\midrule
scene 1 & 2 & 100\% & 2 & 100\% \\
scene 2 & 2 & 100\% & 2 & 100\%\\ 
scene 3 & 2 & 100\% & 2 & 100\% \\ 
scene 4 & 4 & 80\% & 2 & 100\% \\
scene 5 & 6 & 100\% & 6 & 100\% \\ 
\bottomrule
\end{tabular}
\label{tab:use_dog_vs_human}
\end{table}

\begin{table}[t!]
\centering
\caption{Collaboration comparison of G1-Go2 system vs. human baseline}
\small
\begin{tabular}{c ccc ccc}
\toprule
\multirow{2}{*}{} & \multicolumn{3}{c}{G1-Go2 System} & \multicolumn{3}{c}{Human baseline} \\ 
\cmidrule(lr){2-4} \cmidrule(lr){5-7}
 & $N_E$ & $V_{GE}$$\downarrow$ & $N_\mathrm{move}$$\downarrow$ & $N_E$ & $V_{GE}$$\downarrow$ & $N_\mathrm{move}$$\downarrow$ \\ 
\midrule
scene 1 & 2 & 0.98 & 5 & 2 & 0.98 & 2 \\
scene 2 & 1 & 0.86 & 2 & 1 & 0.91 & 2\\ 
scene 3 & 2 & 0.76 & 5 & 1 & 0.91 & 2\\ 
scene 4 & 5 & 0.72 & 6 & 4 & 0.88 & 5\\
scene 5 & 6 & 0.96 & 9 & 5 & 1.03 & 6 \\ 
\bottomrule
\end{tabular}
\label{tab:coordination_vs_human}
\end{table}

\begin{table*}[t!]
\centering
\caption{Comparative Ablation Study on Heterogeneous Agents}
\small
\begin{tabular}{c cccc cccc}
\toprule
\multirow{2}{*}{} & \multicolumn{4}{c}{G1-Go2} & \multicolumn{4}{c}{G1-only} \\ 
\cmidrule(lr){2-5} \cmidrule(lr){6-9}
 & TIME$\downarrow$ & PS$\uparrow$ & CR$\uparrow$ & RSME$\downarrow$ & TIME$\downarrow$ & PS$\uparrow$ & CR$\uparrow$ & RSME$\downarrow$ \\ 
\midrule
scene 1 & 64.00 & 68.18 & 100\% & 85.99 & 43.12 & 67.47 & 100\% & 86.69\\ 
scene 2 & 18.22 & 98.08 & 100\% & 14.42 & 16.14 & 75.59 & 100\% & 124.12 \\ 
scene 3 & 28.58 & 96.74 & 100\% & 40.75 & N/A & N/A & 33.33\% & N/A \\
scene 4 & 120.58 & 92.35 & 100\% & 120.91 & N/A & N/A & 40.00\% & N/A \\
\bottomrule
\end{tabular}
\label{tab:ablaG1}
\end{table*}

In this section, we conduct a controlled comparative evaluation between TZPP and human operators on identical navigation tasks. To ensure fairness, two naive participants were recruited and provided with the same perceptual inputs as the robotic system, namely real-time first-person visual streams from both the humanoid and quadruped sensors. No global map or third-person view was available. In addition, the control interface and action constraints were kept consistent with those of the autonomous system, ensuring that both human operators and TZPP operated under equivalent information and execution conditions.

\begin{table}[t!]
\centering
\caption{Open area comparison of G1-Go2(without Mode X), \\ G1-Go2 system and human baseline}
\small
\begin{tabular}{c ccccc}
\toprule
 & Time$\downarrow$ & $CCR_q$$\uparrow$ & $N_1$$\downarrow$ & $N_2$$\downarrow$ & $N_3$$\downarrow$ \\ 
\midrule
G1(no X)-Go2 & 43.78 & 56.27\% & 0.6 & 1.4 & 3  \\
G1-Go2 & 46.28 & 86.00\% & 0 & 0.2 & 0.6 \\ 
Human & 54.60 & 80.00\% & 0 & 0 & 0.2 \\ 
\bottomrule
\end{tabular}
\begin{tablenotes}
\scriptsize 
\item[1] $N_1, N_2, N_3$ represent $N_\mathrm{rev}^h$, $N_\mathrm{rev}^q$, $N_\mathrm{rot}^q$ respectively.
\end{tablenotes}
\label{tab:ablaX_vs_human}
\end{table}

\begin{table}[t!]
\centering
\caption{Obstacle Handling comparison of G1-Go2(without Mode Y), \\ G1-Go2 system and human baseline}
\small
\begin{tabular}{c c c}
\toprule
 & Time$\downarrow$ & $V_\mathrm{avoid}$  \\
\midrule
G1(no Y)-Go2 System & N/A & 0.23 \\
G1-Go2 System & 154.89 & 1.00 \\ 
Human baseline & 94.21 & 1.00 \\ 
\bottomrule
\end{tabular}
\label{tab:ablaY_vs_human}
\end{table}

Tables \ref{tab:task_completion_vs_human} to \ref{tab:coordination_vs_human} summarize the results across multiple metrics. Overall, the G1-Go2 system achieves performance comparable to human operators in terms of task completion and coordination efficiency. In particular, for the humanoid movement distance ($D$), the autonomous system reaches over $95\%$ of human performance, indicating similar levels of path efficiency under the same perceptual constraints.

Further analysis shows that TZPP maintains stable performance across different scenarios, especially in cases where consistent coordination between exploration and execution is required. However, performance gaps remain in certain complex environments, where human operators may better leverage implicit spatial reasoning or long-horizon planning strategies.

Overall, the results suggest that TZPP can achieve competitive collaborative efficiency under equivalent conditions, while demonstrating consistent adaptability in unseen environments. These findings support the effectiveness of the proposed coordination mechanism without overstating parity with human performance.

\subsection{Ablation Study on Mode X/Y Mechanism}

To evaluate the contribution of the adaptive exploration strategy, we compare the full G1-Go2 system with two degraded variants: one with Mode X disabled (G1-Go2(-X)) and one with Mode Y disabled (G1-Go2(-Y)), under identical task settings. Human operators are included as a reference baseline under the same perceptual and control constraints.

\subsubsection{In Open Scenarios Lacking Landmarks}

The experiments were conducted in open environments without salient visual references (e.g., locating a sofa in a corridor).  Table~\ref{tab:ablaX_vs_human} shows that, disabling Mode X leads to a noticeable decrease in exploration efficiency and task stability. This result suggests that Mode X improves performance in landmark-sparse scenes by reducing repeated exploration caused by visual ambiguity and limited reference cues.

\subsubsection{In Scenarios without Direct Accessible Paths}

The experiments were conducted in obstacle-rich environments requiring detour-based navigation (e.g., navigating around structural barriers or staircases).Table~\ref{tab:ablaY_vs_human} shows that, removing Mode Y significantly reduces success rates and planning efficiency. The results indicate that Mode Y enhances navigation in structurally constrained environments by enabling corridor probing and temporary deviation from the direct goal direction. Overall, the results show that Mode X and Mode Y address different environmental challenges and jointly improve system adaptability across heterogeneous scenarios.

\subsection{Ablation Study on the Necessity of Heterogeneous Multi-Agent Systems for Navigation}

To evaluate the role of heterogeneous collaboration, we compare the full G1-Go2 system with a single-agent G1 setup under identical task conditions. As shown in Table~\ref{tab:ablaG1}, the heterogeneous system achieves consistently better performance in terms of task completion rate, path efficiency, and navigation stability.This improvement stems from the complementary capabilities of the two agents. The quadruped enhances exploration by providing wider perceptual coverage and accessing regions that are difficult for the humanoid, while the humanoid focuses on task coordination and goal-directed execution. This division of roles reduces inefficient exploration and improves robustness, particularly in scenarios where the target is not directly observable or requires detour-based navigation. These results highlight the practical importance of embodiment complementarity in heterogeneous multi-agent systems.

\section{CONCLUSIONS}

This paper presented Triple Zero Path Planning (TZPP), a heterogeneous multi-agent navigation framework that operates without task-specific training, prior environmental knowledge, or simulation pre-adaptation. By leveraging a coordinator–explorer architecture between a humanoid and a quadruped robot, the system exploits embodiment complementarity to decouple exploration and execution, enabling effective navigation in previously unseen real-world environments.

To tackle challenges in landmark-sparse and non-directly accessible environments, we propose an adaptive exploration strategy with Mode X and Mode Y. Mode X enhances efficiency in open, visually sparse scenes, while Mode Y supports detour-based navigation in structurally constrained settings. Extensive experiments, including ablations and controlled comparisons with human operators, show that the system achieves near-human path efficiency and consistent performance, with both modes contributing significantly by addressing complementary environmental challenges.

Overall, TZPP provides a practical coordination mechanism for heterogeneous robotic systems and highlights the importance of embodiment complementarity for robust navigation. The proposed framework offers a scalable design perspective for deploying multi-agent systems in dynamic and structurally complex environments.

\bibliographystyle{ieeetr}

\end{document}